# Agent Smith: Machine Teaching for Building Question Answering Agents

Ashok Goel[1], Harshvardhan Sikka[1] and Eric Gregori[1]

[1]*Design and Intelligence Lab, School of Interactive Computing, Georgia Institute of Technology, North Ave NW, Atlanta, GA 30332*

**Abstract**
Building AI agents can be costly. Consider a question answering agent such as Jill Watson that automatically answers students' questions on the discussion forums of online classes based on their syllabi and other course materials. Training a Jill on the syllabus of a new online class can take a hundred hours or more. Machine teaching – interactive teaching of an AI agent using synthetic data sets – can reduce the training time because it combines the advantages of knowledge-based AI, machine learning using large data sets, and interactive human-in-loop training. We describe Agent Smith, an interactive machine teaching agent that reduces the time taken to train a Jill for a new online class by an order of magnitude.

## 1. Introduction

Intelligent systems, and virtual assistants in particular, have proliferated in their utility and application across a variety of domains [1]. This includes online education and learning, a domain that has rapidly gained in popularity over the past decade. The unprecedented scale of online education, and other domains including remote work, has resulted in a variety of new problems ripe for AI to address. For example, a frequent criticism of online learning environments is their notably low retention rates [2] and lower student satisfaction [3]. Another issue with these learning environments is the lack of community resulting from the absence of face to face interactions among learners [4, 5]. Virtual assistants provide unique approaches and solutions to many of the issues referenced above. When considering the sheer quantity of problems that present themselves at the internet scale, there are a staggering number of opportunities that present themselves for virtual assistants. One such virtual assistant, the Jill Watson Q&A system, aims to enhance and support instructor and learners in online learning environments through automatic question answering, thereby enhancing teacher presence[6, 7].

A key challenge in the development of virtual assistants aiming to solve the issues mentioned above, and intelligent systems more generally, is designing, training, and configuring them to be effective in new contexts [8, 9]. This requires the involvement of a domain expert to appropriately serve as an teacher, effectively "teaching" the machine learning system to solve its target task in a given domain. However, teaching intelligent systems and virtual assistants can be a costly and difficult process, often resulting in an enormous amount of person hours required on the part of the domain expert [10]. Machine Teaching is a collection of approaches explicitly aimed at solving the difficulties that lie in enabling domain experts to effectively teach machine learning systems [10, 8, 9, 11, 12, 13]. In developing easy and fast approaches to teaching machines, the space of potential opportunities across a wide variety of domains opens up.

In this paper, we present Agent Smith, an interactive machine teaching environment that helps rapidly train custom Jill Watson Q&A agents in new domains, saving domain experts hundreds of hours in "teaching" time. First, we explore the capabilities of Jill Watson agents in online learning environments. Then, we outline the role, process, and underlying mechanisms in the Agent Smith system. Following this, we demonstrate the Agent Smith[1] process in an example classroom scenario, moving from mapping the domain of classroom information to a machine readable format through template question formation and dataset generation, and finally classification model training. Subsequently, we present empirical evidence for the success of Agent Smith in its objective of saving users time when training a new Jill Watson system, and statistics about its deployments. Finally, we discuss the implications of this approach as well as related work and future directions.

## 2. Question Answering in Jill Watson

The Jill Watson Q&A agent aims to enhance and support instructor and learners in online learning environments. It accomplishes this through facilitating routine question answering reducing the associated burden on the part of the course instructor. This in turn allows instructors to leverage their time and presence more effectively in large scale learning environments with many students. As a result, instructors can reach more learners effectively and spend more time engaging with learners. For learners, Jill Watson provides enhanced learning assistance, increasing learner engagement in online learning environments by answering questions quickly and effectively.

The current Jill Watson agent is the result of several iterations of design, development, and evaluation mostly in the context of the Online Master of Science in Computer Science (OMSCS, https://omscs.gatech.edu) program at Georgia Tech, which was launched in 2014. In the OMSCS setting, students would interact with the professor and teaching assistants through online discussion forums like Piazza, as well as through learning management systems (LMS) like Canvas. In the early days of the OMSCS, the problem of teacher presence mitigation was clearly identified. For example, during the first deployment of the online course on Knowledge Based

---

[1]The name "Agent Smith" is inspired by the character with the same name in the film "the Matrix", an agent who could clone himself. Here the machine teaching process using Agent Smith creates, or "clones", Jill Watson for each new context.

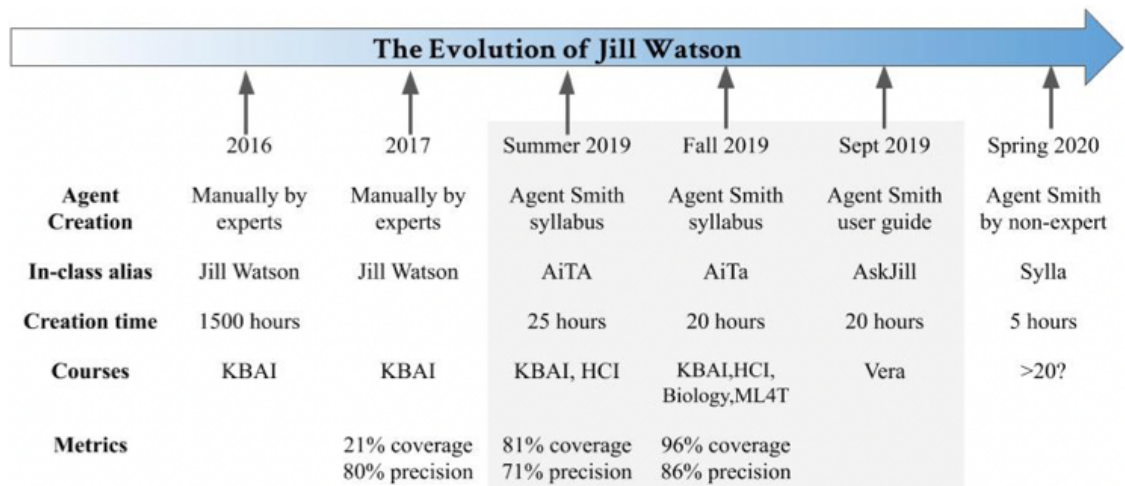

**Figure 1:** The iterative development of Jill Watson through time, broken down into several characteristics, including its alias during classroom deployment, the method behind its creation, the courses it was deployed in, the total time taken to create a Jill Watson agent, and its coverage and precision metrics.

AI (KBAI, https://omscs.gatech.edu/cs-7637-knowledge-based-artificial-intelligence-cognitive-systems), students posted hundreds of questions on the associated discussion forums, immediately resulting in difficulties for the instruction team to successfully answer them all. This prompted the development of the first version of Jill Watson, a virtual teaching assistant named Jill Watson. The goal behind Jill Watson was similar to the recent version of Jill Watson: automatically answering questions that popped up frequently related to core syllabus information. This version of Jill Watson, referred to as Jill Watson 2016, used a collection of question and answer pairs organized into various question categories in a preliminary grouping. Following this, the design of Jill Watson was revised and iterated upon, resulting in several important characteristics. Rather than focus on a databank of question and answer pairs, a novel ontology of class syllabi was developed, with different course syllabus information being organized around this ontology for use by the Jill Watson system.

The most recent version of Jill Watson provides increased question answering capabilities across a variety of domains beyond the original online learning scenario. At a high level, Jill Watson uses a hybrid classification approach to separate and answer various questions pertaining to a course syllabus successfully, demonstrated in Figure 2. The first stage in this process uses statistical machine learning methods to categorize incoming questions and label their underlying intents. Following this, a knowledge based classifier parses the question and structures an appropriate response from an underlying knowledge base of relevant course related information.

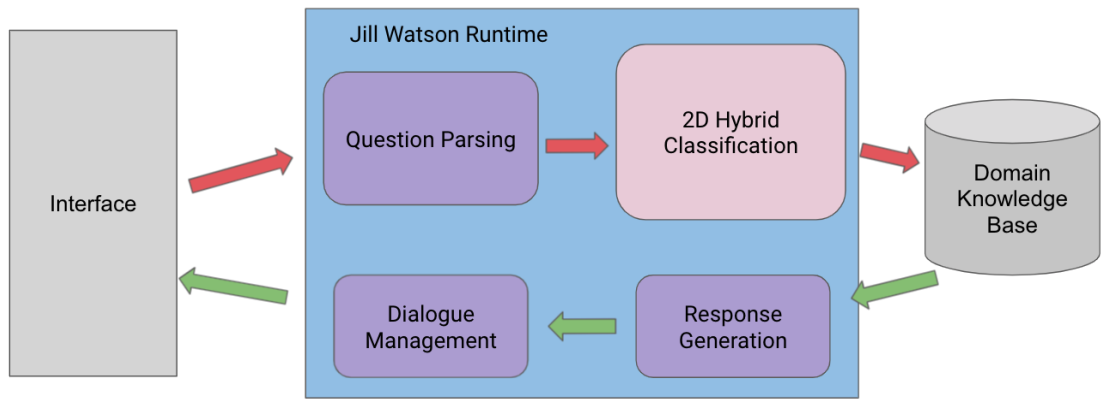

**Figure 2:** High Level Overview of a Jill Watson Agent. Questions are parsed and classified. Following this, responses are generated and submitted back to the user.

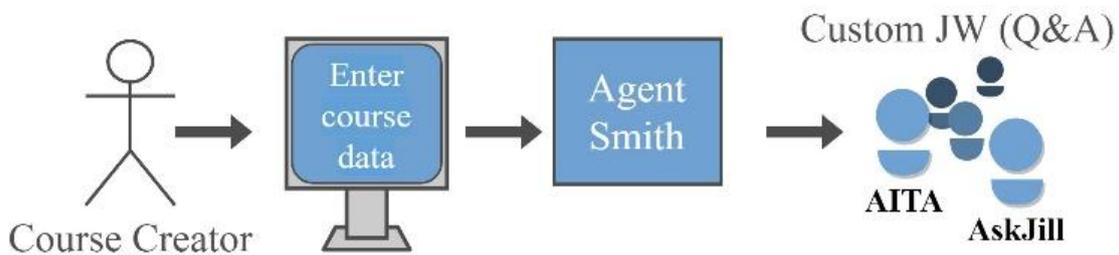

**Figure 3:** Overview of the Agent Smith process and motivation. A Course Creator submits course data and configures the Agent Smith System. Once configured, Agent Smith can train numerous Jill Watson agents for the new domain.

## 3. Machine Teaching in Agent Smith

### 3.1. Agent Smith Overview

Jill Watson Q&A agents are able to support a variety of domains, notably including instructors and learners in online learning environments. However, the utility of developing a new Jill Watson agent for a given class instance or a new domain only makes sense if it is affordable from the perspectives of labor and time associated with each new configuration. Building new Jill Watson agents from unstructured documents like class syllabi or job descriptions often requires the time and labor of an expert in the domain area, which is inherently costly in many scenarios.

To resolve this constraint, we introduce Agent Smith, an interactive machine teaching environment for developing Jill Watson agents in new domains. Agent Smith has the explicit goal of making it easy and efficient for domain experts to generate custom Jill Watson systems.

Agent Smith uses an knowledge-based process similar to the one used by researchers and domain experts when configuring a new Jill Watson instance. At a high level, the Agent Smith environment works by operating on two distinct inputs. The first is the aforementioned ontology that defines the domain or the application area, referred to generally as the Knowledge Base (KB). Conceptually, the Agent Smith generator uses a form for events (When is an assignment due?) and semantic memory for facts and concepts (What is the course late work policy?). The KB contains concepts and relations that form the body of the knowledge that an Jill Watson Agent will seek to explain. The other important input to the Agent Smith system is a set of question templates that capture the structural form of questions asked by potential users in the domain. In the course context, these general templates may capture questions related to deadlines, key terms, topics covered in different areas of the course, etc.

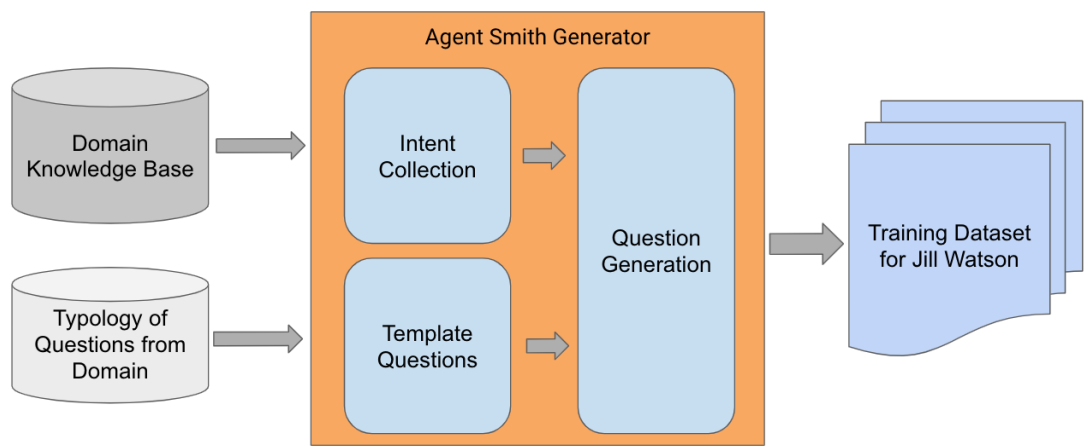

**Figure 4:** Agent Smith Architecture Overview

Agent Smith combines these two knowledge bases as part of a combinatorial generation process to create large sets of questions and their associated intents. A visual overview of this process is given in Figure 4. These question-intent pairs can then be used to train the machine learning classifier used in a Jill Watson agent. The overall result of the Agent Smith system is a much faster turnaround time in training a new Jill Watson agent and a systematic approach to achieving a high degree of question coverage for a target domain. Importantly, using Agent Smith, the course creator can create an agent in less than 20 hours, saving the course creator over 100 hours in answering questions a typical online virtual classroom. In the next section, we walk through an example of Agent Smith in action, demonstrating how the agent smith process results in a trained Jill Watson agent for a specific class.

### 3.2. Agent Smith in Action

Agent Smith has been applied to building Jill Watson Agents in several courses as part of the OMSCS program at the Georgia Institute of Technology. Here, we walk through the Agent Smith process to build a Jill Watson agent for the Knowledge Based Artificial Intelligence course taught in the OMSCS program during the summer of 2020, demonstrating the knowledge bases and the process behind their construction, the template questions and the process behind their creation, the generative process and final dataset, and the trained classifier.

The first step in enabling the Agent Smith system to rapidly train Jill Watson Q&A agents is to map the course domain to the unstructured and structured databases mentioned earlier. This is done through reviewing examples of the domain that is to be defined and creating a taxonomy based on similarities in the domain examples. Categories in the taxonomy are assigned a label for the purpose of associating potential training questions with a response, where the focus is that each label is unique. An example excerpt of a high level taxonomy for courses is given in Figure 5.

| Unstructured Taxonomy | Structured Taxonomy |
|---|---|
| coursedescription | description |
| teachingstaff | weight |
| officehours | releasedate |
| learning | duedate |
| lateworkpolicy | week |
| intellectualpropertypolicy | submission |
| importantdates | grading |
| disabilityaccomodations | duration |
| courseprerequisites | estimatedtime |
| coursematerials | url |
| grade | resources |

**Figure 5:** Example taxonomy of categories in an online course, divided into unstructured and structured. These are categories used in Agent Smith during training of Jill Watson agents for OMSCS courses, including Knowledge Based Artificial Intelligence.

Following the creation of a taxonomy and its associated categories, structured entities and unstructured categories can be identified, defined by the granularity and structure of the associated information. These categories and their associated information are mapped to the structured and unstructured databases. We include the schemas in Figure 7. The structured table captures entities explicitly specified in its schema, while the unstructured table aims to capture information that does not hold explicit identified entities, and presents more general information. In practice, this means the structured table has a set of specific pieces of information associated with each category captured in its schema, while the unstructured table contains unstructured text responses associated with the category. These tables are the same tables used by the Jill Watson agent during runtime when answering questions.

Following the mapping of the domain to the unstructured and structured knowledge bases described above, the next important step is to collect a set of questions that will likely be asked about a given domain, the Knowledge Based AI class in our example. These questions are then mapped to a general list of template questions and their associated intents. Template questions seek to capture general forms of questions that occur within categories in the domain, representing a large set of potential questions. These templates were manually designed by extracting common patterns in corpuses of student question data gathered from previous deployments of Jill Watson in educational courses in the OMSCS. Question analysis approaches, including the frequency of different n-gram occurrences in the dataset is used to highlight patterns in questions and the objects they pertain to. A high level overview of this process can be seen in Figure 6. This is a preliminary approach, and we are exploring using large natural language processing models trained on general text data to augment template design in future work. An example schema for the database of template questions is given in Figure 7.

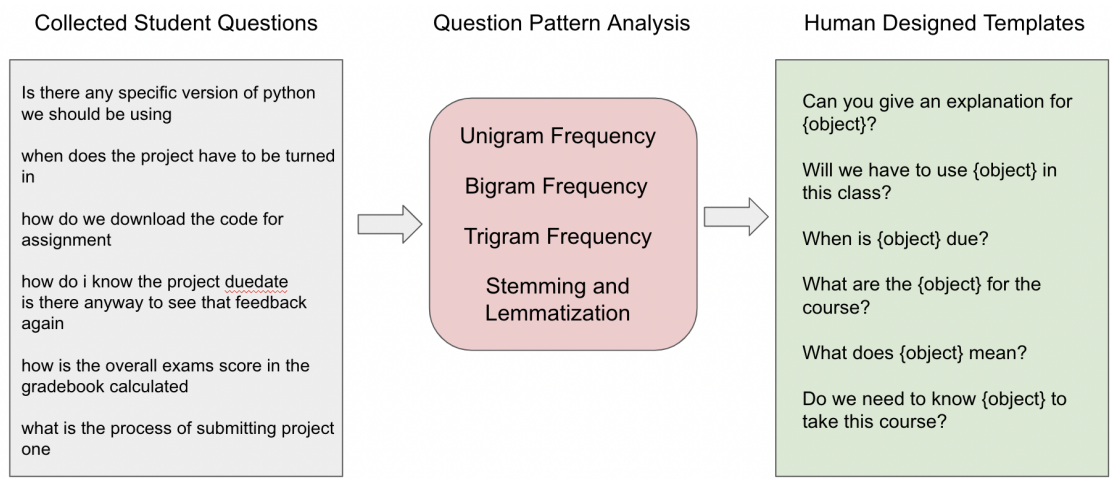

**Figure 6:** Conceptual overview of template design process. Starting with real user questions, various analytical approaches to understanding common patterns in question structure are used to ultimately design templates. Example template and intent associations can be observed in Table 1.

In Table 1, we demonstrate example template questions for a subset of categories in the domain, along with the number of total templates per category. We can see that different categories hold a different number of question templates as there are varied structural forms present in those categories as compared to others. For example, the "coursematerials" category contains 85 template variations, while "teachingstaff" only contains 9.

With the creation of the domain knowledge base, in both its structured and unstructured forms, as well as the template questions database, Agent Smith can be used to create a large dataset of example question answer pairs to train a Jill Watson Q&A agent. When used, the Agent Smith system combinatorially combines template questions with various structured and unstructured keywords that represent the categories in both sections of the domain database. The result is a dataset of questions orders of magnitude larger than previously hand collected and labeled datasets for the KBAI course, in a fraction of the time. In Table 2, we demonstrate a few examples of example questions generated from some of the templates demonstrated in the previous table, along with their associated intents.

Following the running of Agent Smith on the KBAI knowledge bases and template question dataset created from the Fall 2020 version of the class, 24,587 questions and intent pairs were created and subsequently used to train the machine learning classifier underlying the 2D

| **Unstructured Table** | |
|---|---|
| id | int(11) |
| label | text |
| keywords | text |
| responsetext | text |
| responsesource | text |

| **Template Question Table** | |
|---|---|
| id | int(11) |
| keyword_source | text |
| template | text |
| example | int |

| **Structured Table** | |
|---|---|
| id | int(11) |
| identified | text |
| object_keywords | text |
| description | text |
| weight | text |
| releasedate | text |
| duedate | text |
| ... | ... |
| estimatedtime | text |
| url | text |
| resources | text |
| guideline | text |

**Figure 7:** Type of knowledge that knowledgebases used by Agent Smith to create training datasets. Note that for brevity, some categories were skipped in the Structured Table.

classification portion of the Jill Watson Q&A agent.

The core focus of Agent Smith is to reduce the time cost of developing a new Jill Watson Agent in a domain. We estimate the time cost for creating, training, and configuring Jill Watson was around 500 person hours in 2016. Through the use of the Agent Smith machine teaching system, the estimated time cost of the development of a new Jill Watson agent in a given domain is around 25 person hours, and can be less. Notably, Agent Smith has been deployed and used in more than 20 course deployments over the past few years in the OMSCS program, and has also enabled the creation of Jill Watson style agents in other domains.

## 4. Discussion

### 4.1. Related Work

Agent Smith draws from ideas in both the field of data augmentation and the field of machine teaching. Data augmentation is an area focusing on methods of increasing training data with the need to go and collect more data [14]. Data Augmentation, and even Data Augmentation approaches to Natural Language Processing [15], covers a large variety of approaches. In this section we focus on related work in Data Augmentation in the area of rule based approaches and data augmentation techniques used for Q and A applications. Easy Data Augmentation, or EDA, proposes using random perturbation operations to include random insertion, swaps, and deletions [16]. Linguistic knowledge based approaches have also been proposed, incorporating prior knowledge to form the augmentation, and templates have been used successfully in Q&A applications prior to this as well [17, 18]. Our approach makes use of manually designed templates by educational technology researchers. This is appropriate for the generation of Jill Watson agents due to the niche nature of the educational context as opposed to the general language domain that other approaches may operate in.

Table 1
Training Templates

| Categories | Example Training Templates | Total Templates |
|---|---|---|
| coursedescription | Will we learn about user in this class? | 27 |
| teachingstaff | Who teaches this class? | 9 |
| officehours | When are office hours this week? | 13 |
| lateworkpolicy | What is the penalty for submitting work past the deadline? | 31 |
| intellectualpropertypolicy | Can I post my work on a public platform? | 11 |
| importantdates | When is the object? | 8 |
| disabilityaccomodations | Where can I find information about Disability Services? | 20 |
| learning | What are the learning goals of this class? | 47 |
| courseprerequisites | Do we need to know object to take this course? | 18 |
| coursematerials | What are the object for the course? | 85 |
| definition | Can you give an explanation for object? | 14 |

Table 2
Generated Question Examples

| Example Intent | Generated Question |
|---|---|
| coursedescription | Will we learn about artificial intelligence in this class? |
| importantdates | When is the withdraw date? <br> When is the start date? |
| courseprerequisites | Do we need to know python to take this course? <br> Do we need to know C to take this course? |
| coursematerials | What are the course materials for the course? <br> What are the websites for the course? |

The field of Machine Teaching is more focused, and includes a few notable works. Simard et al propose a perspective on machine teaching that focuses on improving the efficacy of teachers, putting a strong emphasis on the interaction of a teacher with the data [8]. Most notably, they advocate for interfaces that decouple the knowledge related to a particular machine learning system from the interaction of a teacher, allowing domain experts not familiar with machine learning systems to participate in teaching these systems. Zhu et al present machine teaching in a more formal setting, characterizing a problem space that describes machine teaching and all associated problems [11, 9]. Our approach focuses on combining knowledge-based AI and machine learning techniques for designing a machine teaching approach and associated processes for rapid teaching of hybrid machine learning systems.

### 4.2. Future Work

Going forward, there are several exciting areas of research in applying the Agent Smith system, as well as Machine Teaching approaches in general. Enabling Machine Teaching systems to teach explanatory agents to understand not just static, structured information, but also dynamic information will allow for the rapid design and training of agents that can lend understanding about the state of Software Systems. In particular, we seek to design Agent Smith such that it can make use of knowledge of design of an AI agent including its internal states to build Jill Watson Q&A agents for explaining the functioning of the AI agent. Another notable area is to include better teaching interfaces and data augmentation approaches that allow domain experts to more easily configure an Agent Smith system to build Q&A agents.

## 5. Conclusion

In this paper, we introduce a machine teaching system for the rapid training of Jill Watson Q & A agents in new online course settings. In particular, we demonstrate the process of mapping a new course domain to a structured knowledge base, and explain how a newly developed machine teaching approach, dubbed Agent Smith, can use this information to generate large data sets of example questions that cover much of what a student may ask in these course settings. We also demonstrate an example of Agent Smith in action, moving from the mapping of a domain through to a trained agent in an example graduate course at the Georgia Institute of Technology. Applying machine teaching to virtual assistants opens up an exciting set of directions, including new interfaces rapidly building virtual assistants that understand and model software systems, two areas which we intend on investigating in the future.

### 5.1. Acknowledgements

We are grateful to IBM for providing us with access to its Watson platform. Jill Watson and Agent Smith have been developed at Georgia Tech independently. We thank the Jill Watson team at Georgia Tech Design Intelligence Laboratory including Vrinda Nandan, Spencer Rugaber, and Karan Taneja for many discussions about Agent Smith. We are grateful to Georgia Tech for supporting this research through internal seed grants.

question answering, arXiv preprint arXiv:2004.10157 (2020).